# A Long Short-term Memory Based Recurrent Neural Network for Interventional MRI Reconstruction

Ruiyang Zhao, Zhao He, Tao Wang, Suhao Qiu, Pawel Herman, Yanle Hu, Chencheng Zhang, Dinggang Shen, Bomin Sun, Guang-Zhong Yang, and Yuan Feng

**Abstract—Interventional magnetic resonance imaging (i-MRI) for surgical guidance could help visualize the interventional process such as deep brain stimulation (DBS), improving the surgery performance and patient outcome. Different from retrospective reconstruction in conventional dynamic imaging, i-MRI for DBS has to acquire and reconstruct the interventional images sequentially online. Here we proposed a convolutional long short-term memory (Conv-LSTM) based recurrent neural network (RNN), or ConvLR, to reconstruct interventional images with golden-angle radial sampling. By using an initializer and Conv-LSTM blocks, the priors from the pre-operative reference image and intra-operative frames were exploited for reconstructing the current frame. Data consistency for radial sampling was implemented by a soft-projection method. To improve the reconstruction accuracy, an adversarial learning strategy was adopted. A set of interventional images based on the pre-operative and post-operative MR images were simulated for algorithm validation. Results showed with only 10 radial spokes, ConvLR provided the best performance compared with state-of-the-art methods, giving an acceleration up to 40 folds. The proposed algorithm has the potential to achieve real-time i-MRI for DBS and can be used for general purpose MR-guided intervention.**

## I. Introduction

D EEP brain stimulation (DBS) involves a neural intervention procedure where electrodes are inserted into specific nuclei for the treatment of neurological disorders such as Parkinson's disease [1], [2]. To make sure the electrodes reach the target nucleus, current practice in DBS usually uses pre-operative CT and MR images for surgical planning, but intraoperative guidance is still lacking. This could lead to an offset from the target position, due to the brain shift caused by craniotomy, posing a potential risk for surgical performance or even misplacement of the electrodes [3], [4]. Therefore, it is in great desire to have real-time visualization of the interventional process that could provide visual feedback for neurosurgeons.

Interventional Magnetic Resonance Imaging (i-MRI) is a non-invasive imaging technique, which provides image guidance for therapeutic procedures [5], [6]. The soft tissue contrast and accurate discrimination between normal and abnormal soft tissues provided i-MRI unique advantages to other imaging methods. A fast and robust i-MRI technique that can help track the electrode position during the DBS surgery will significantly improve the treatment accuracy and patient outcome [7], [8], [9].

To accelerate MRI acquisition, classical fast MRI techniques include balanced steady-state free precession (bSSFP) [10], [11], keyhole imaging [12], [13], and parallel imaging [14], [15]. Despite that these methods were computationally efficient, they could not provide enough image resolution for the interventional features and fine structures of the brain. In the past decades, compressed sensing (CS) has drawn significant attention from the fast MRI community [16], [17], [18]. Recently, we have proposed a low-rank and framelet-based sparsity decomposition algorithm and demonstrated its potential for i-MRI reconstruction [19]. However, reconstruction with these iterative CS-based methods requires extra computational time that impedes the implementation in clinical scenarios [20]. Therefore, a method that is suitable for real-time image guidance of intervention is still needed.

Recent development in deep learning (DL) has demonstrated

The work was supported in part by the National Natural Science Foundation of China under Grant 31870941, in part by the Natural Science Foundation of Shanghai under Grant 22ZR1429600, in part by the Science and Technology Commission of Shanghai Municipality under Grant 19441907700, and in part by SJTU Global Strategic Partnership Fund (2019 SJTU-KTH) under Grant WF610561702/067. (*Ruiyang Zhao, Zhao He, and Tao Wang contributed equally to this work.*) (*Corresponding author: Yuan Feng.*)

R. Zhao*, Z. He, S. Qiu, and Y. Feng are with School of Biomedical Engineering, Shanghai Jiao Tong University, Shanghai 200240, China. (e-mail: zhzry@163.com; sjtu_hz_0028@sjtu.edu.cn; qiusuhao@sjtu.edu.cn; fengyuan@sjtu.edu.cn.) *R. Zhao has graduated and is now with the Department of Electrical and Computer Engineering, University of Illinois Urbana-Champaign, Urbana, IL, United States.

T. Wang, C. Zhang, and B. Sun are with the Department of Neurosurgery, Center for Functional Neurosurgery, Ruijin Hospital affiliated to Shanghai Jiao Tong University School of Medicine, Shanghai, China. (e-mail: wtao_sjtu@sjtu.edu.cn; i@cczhang.com; bomin_sun@163.com).

P. Herman is with the Division of computational science and technology, KTH Royal Institute of Technology, Stockholm, SWEDEN (e-mail: paherman@kth.se).

Y. Hu is with the Department of Radiation Oncology, Mayo Clinic in Arizona, Phoenix, AZ, USA (e-mail: Hu.Yanle@mayo.edu).

D. Shen is with the School of Biomedical Engineering, ShanghaiTech University, Shanghai 201210, China (e-mail: dgshen@shanghaitech.edu.cn).

G. Yang is with the Institute of Medical Robotics, Shanghai Jiao Tong University, Shanghai 200240, China (e-mail: gzyang@sjtu.edu.cn)

that a deep network is an effective way to perform fast and accurate image reconstruction [21], [22], [23]. J. Schlemper et al. proposed a cascade convolutional neural network to reconstruct dynamic MR images [24]. G. Yang et al. proposed a deep de-aliasing generative adversarial network (DAGAN) for fast compressed sening MRI reconstruction [25]. In addition, networks that unroll the optimization process of the objective function were proposed for fast MRI reconstruction. Typical networks include ISTA-Net [26], ADMM-CSNet [27], SLR-Net [28], and L+S-Net [29]. To utilize the temporal information of the image series during the dynamic process, a recurrent neural network (RNN) was successfully used for the reconstruction of dynamic cardiac [30], [31], and abdomen [32] images in an end-to-end fashion. Convolutional Long Short-term Memory (Conv-LSTM), as a powerful model of RNN architectures [33], has been used for dynamic imaging [34], [35], [36]. Although these methods show great performance for the reconstruction of dynamic imaging, they were not specifically tailored for i-MRI.

In this study, we proposed a Conv-LSTM based RNN for i-MRI reconstruction. The temporal image information during the intervention was exploited by the Conv-LSTM block. A fully sampled preoperative reference image was used for RNN initialization. Datasets were generated from the pre-operative and post-operative MR images for training and testing. Reconstruction results were compared with those from CS and DL-based methods.

## II. Problem Formulation

### A. CS-MRI

Let $\mathbf{x} \in \mathbb{C}^D$ represent a sequence of interventional images to be reconstructed, where $D = N_x N_y N_t$. Each image consists of $N_x N_y$ pixels and $N_t$ denotes the number of frames. In a radially sampled k-space, the acquired data $\mathbf{y} \in \mathbb{C}^P$, where $P = N_p N_s N_t$. The k-space contains $N_s$ spokes sampled with $N_p$ points. Typically, taking $N_s = 402$ would give a fully sampled image satisfying Nyquist criteria (Fig. 1) with image size 256×256.

The problem can be formulated by reconstructing the interventional MR images $\mathbf{x}$ from the acquired k-space data $\mathbf{y}$:

$$\mathbf{y} = \mathbf{E}\mathbf{x} + \mathbf{err}, \tag{1}$$

where $\mathbf{y} \in \mathbb{C}^P (P \ll D)$, $\mathbf{E}$ is the encoding matrix and $\mathbf{err} \in \mathbb{C}^P$ is the acquisition error. In the radial sampling scheme, $\mathbf{E} = \mathbf{M}\mathbf{P}\mathbf{F}$, where $\mathbf{F} \in \mathbb{C}^{N \times P}$ is the Fourier transform, $\mathbf{P} \in \mathbb{C}^{P \times N}$ is the projection acquisition in k space, and $\mathbf{M} \in \mathbb{C}^{P \times P}$ is the undersampling mask selecting certain projection lines for each frame. Reconstructing the undersampled images is an ill-posed problem that can be solved by an unconstrained optimization problem.

$$\underset{x}{\mathrm{argmin}} \parallel \mathbf{E}\mathbf{x} - \mathbf{y} \parallel_2^2 + \lambda R(\mathbf{x}), \tag{2}$$

where $\parallel \mathbf{E}\mathbf{x} - \mathbf{y} \parallel_2^2$ is a data fidelity term, and $R(\mathbf{x})$ is a regularization term typically employed as the $l_0$ or $l_1$ norm in the sparsifying domain of $\mathbf{x}$, and $\lambda$ is a regularization parameter.

### B. Deep Learning-Based Interventional CS-MRI

Different from dynamic image reconstruction that can be carried out retrospectively after acquiring all imaging data, real-time i-MRI requires reconstructing image frames online. The time frame determines the actual temporal resolution of i-MRI. Inspired by the prior information-based technique using a reference frame [37], we can also treat the image before intervention as a reference prior to reconstruction. To improve the reconstruction performance and effectively utilize the temporal coherence during the interventional process, we introduced an RNN network to incorporate the reference image and consecutive interventional frames.

$$\mathbf{x_{rnn}} = f_{rnn}(\mathbf{x_u}|\theta, \mathbf{x_{ref}}), \tag{3}$$

where $\mathbf{x_u}$ represents a series of reconstructed images from radially undersampled k-space data, $\boldsymbol{\theta}$ is the meta parameter of the RNN, $\mathbf{x_{ref}}$ is the reference image and $\mathbf{x_{rnn}}$ denotes the reconstructed images from the RNN output. By introducing the $\mathbf{x_{rnn}}$ into the regularization term, the reconstructed final image, $\mathbf{x}$, can be obtained by solving the following problem:

$$\underset{x}{\mathrm{argmin}} \parallel \mathbf{E}\mathbf{x} - \mathbf{y} \parallel_2^2 + \lambda \parallel \mathbf{x} - \mathbf{x_{rnn}} \parallel_2^2. \tag{4}$$

The fully sampled reference image $\mathbf{x_{ref}}$ was acquired without interventional features and was used as an initializer of the proposed method.

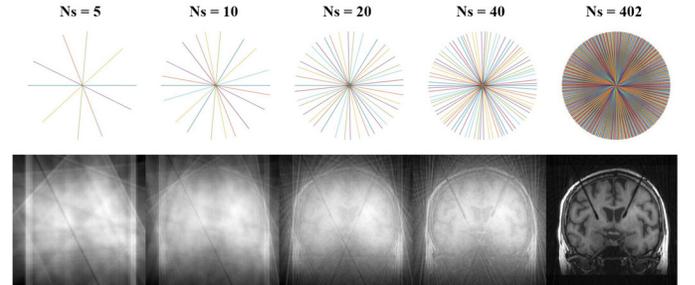

Fig. 1. Typical interventional MR images in DBS surgery with different sampling spokes and the corresponding NUFFT reconstruction results.

## III. Methods and Materials

### A. Data Acquisition and Preparation

In this study, we focused on the scenario of brain i-MRI in DBS surgeries. The interventional feature is a cannula in which the electrode will be passed through to reach the target nucleus inside the brain. We acquired image data from a total of 29 patients who received a DBS surgery. For each patient, 3D whole-brain MRI data (before and after surgery) were collected and used for training and testing dataset preparation (Fig. 2).

Interventional features including implanted electrodes and wires were first extracted from the post-surgery images and then used for intraoperative intervention image $\mathbf{x}_t^I$ simulation. At a different time point t, the simulated intervention feature changes position along the intervention direction. To make sure the model can robustly capture the arbitrary movement of the interventional feature, for consecutive five frames, the position of intervention feature changes through data augmentation including randomly rotate and shift images, 2400 slices (5 frames for each slice) generated from 23 patients were used to construct the training and validation data sets. A total of 188 slices (5 frames for each slice) from another 6 patients were used to construct the test.

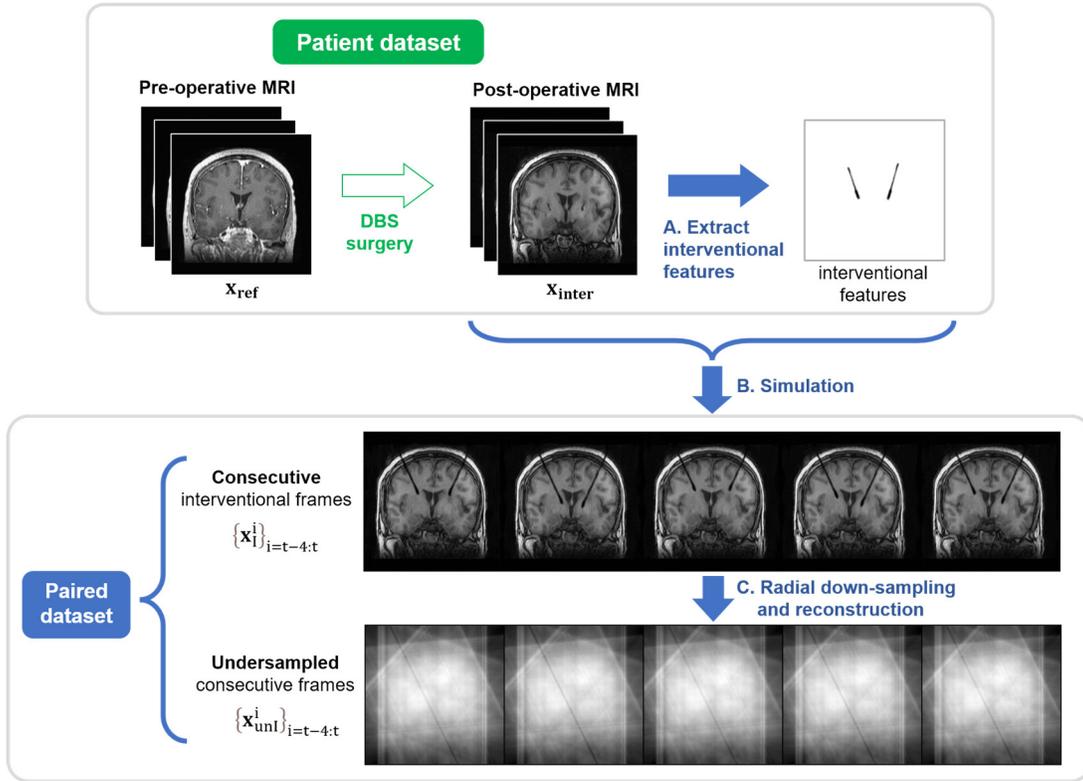

Fig. 2. An overview of the data processing pipeline (Using 5 frames as an example). For each patient, reference image $\mathbf{x}_{ref}$ and post-surgery image $\mathbf{x}_{inter}$ are firstly collected. A: Extracting interventional features from the post-surgery image $\mathbf{x}_{inter}$; B: Generating simulated interventional images $\mathbf{x}_i^t$ at different time $t$ with the extracted interventional features; C: Performing golden angle radial trajectory down-sampling; Paired datasets are prepared for training and validation.

We adopted a golden angle radial sampling strategy for k-space sampling. For each time frame, inputs of a total of 402 sampling radial spokes were collected with 512 readout points for each spoke. For accelerated i-MRI, images were reconstructed with 5, 10, 20, and 40 spokes. The reconstructed image with subsampled k-space data was represented by $\mathbf{x}_{unl}^t$ and was taken as the input of the proposed network.

### B. Network Architecture

The proposed network was composed of one initializer and several RNN blocks (Fig. 3 (a)). To incorporate the prior information from the reference image and utilize the temporal coherence between different frames, inputs of the network consist of two parts: 1) The network was initialized with a fully sampled pre-operative reference image $\mathbf{x}_{ref}$ ; 2) Consecutive frames (Taking five consecutive frames as an example: from previous time point $t-4$ to current time point $t$ ) of undersampled images $\{\mathbf{x}_{unl}^i\}_{i=t-4:t}$ were then fed to the network. The corresponding reconstruction results can be expressed as $\{\mathbf{x}_{rec}^i\}_{i=t-4:t} = f_{rnn}\left(\{\mathbf{x}_{unl}^i\}_{i=t-4:t} \mid \boldsymbol{\theta}, \mathbf{x}_{ref}\right)$.

Each RNN block consists of two cascade CNNs and one data consistency layer (Fig. 3 (b)). To utilize the temporal coherence between different frames, the Conv-LSTM blocks are inserted into CNNs. The Conv-LSTM block determines the current state from the current inputs and previous states, which is an effective way to incorporate information from previous time points. At time $t$ in the $j^{th}$ CNN block, for the $k^{th}$ Conv-LSTM layer $B_{k,j}^t$, cell state $\mathbf{c}_{k,j}^t$ and the hidden state $\mathbf{h}_{k,j}^t$ get updated according to the current input state $\mathbf{x}_{k-1,j}^t$ and the historical states $\mathbf{c}_{k,j}^{t-1}, \mathbf{h}_{k,j}^{t-1}$. The inference process is

$$\mathbf{x}_{k,j}^t = \text{Deconv}(\mathbf{h}_{k-1}), \tag{5}$$

$$\mathbf{c}_{k,j}^t, \mathbf{h}_{k,j}^t = B_k^t(\mathbf{x}_{k,j}^t, \mathbf{c}_{k,j}^{t-1}, \mathbf{h}_{k,j}^{t-1}) \tag{6}$$

Initialized with:

$$\mathbf{x}_1^t = \text{Encoder}(\mathbf{x}_{unl}^t) \tag{7}$$

$$\mathbf{c}_{k,j}^0, \mathbf{h}_{k,j}^0 = \text{Initializer}(\mathbf{x}_{ref}) \tag{8}$$

where Deconv is the deconvolution neural network, and Encoder is the encoding neural network. Initializer takes the reference image as input and provides the initial states $\mathbf{c}_{k,j}^0$, $\mathbf{h}_{k,j}^0$ for LSTM update.

To ensure the data consistency (DC) in k-space, we used a soft-projection method [38] for the data consistency layer, which is designed for the non-Cartesian acquisition scheme. That is, in each iteration, the DC layer serves as a soft projection step and the CNN block plays a role of regularization. The output of the neural network block $\mathbf{x}_{cnn}^k$ is fed into the DC layer, and the output of the data consistency layer $\mathbf{x}^{k+1}$ is then put into the next neural network block for the regularization step. The DC layer is formulated as follows:

$$J_{dc} = \|\mathbf{Ex} - \mathbf{y}\|_2^2 \tag{9}$$

$$\mathbf{x}^{k+1} = \mathbf{x}_{cnn}^k - \alpha\nabla(J_{dc}) = \mathbf{x}_{cnn}^k - \alpha\mathbf{E}^T(\mathbf{Ex}_{cnn}^k - \mathbf{y}) \tag{10}$$

where $J_{dc}$ is the DC term, and $\alpha$ is a learnable parameter.

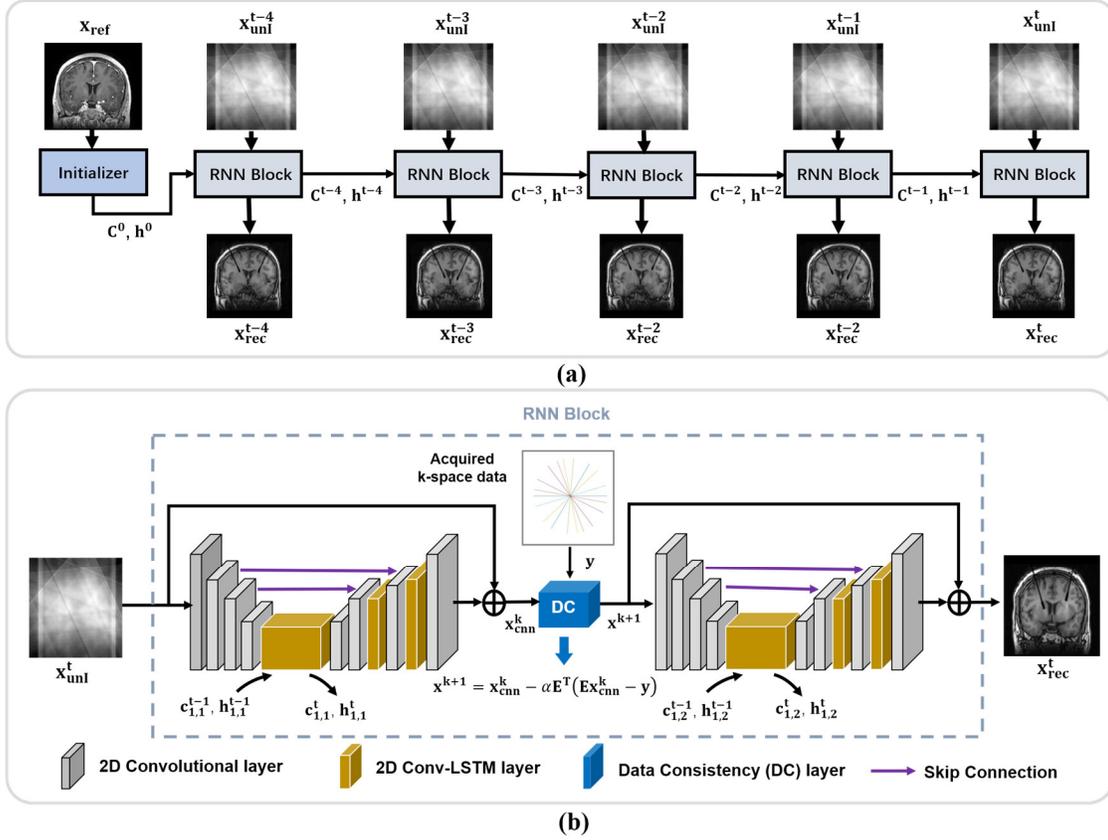

**(a)**

**(b)**

Fig. 3. The architecture of the proposed network. (a) The proposed network consists of one initializer and several RNN blocks. The initializer absorbs reference image, providing prior information for RNN block reconstruction. (b) In each RNN block, the undersampled image $\mathbf{x}_{unl}^{t}$ is reconstructed by two cascade CNNs with one data consistency layer. The Conv-LSTM blocks are inserted in CNN to utilize the temporal coherence between different frames. At the state of $k^{th}$ block in the $j^{th}$ CNN block: $\mathbf{c}_{k,j}^{t}$ and $\mathbf{h}_{k,j}^{t}$ is updated according to the state of $(k-1)^{th}$ Block and the historical states $\mathbf{c}_{k,j}^{t-1}$, $\mathbf{h}_{k,j}^{t-1}$.

Similar to DAGAN [25], the loss function here consisted of four parts: the mean square error loss in the image domain ($L_{iMSE}$), the mean square error loss in the frequency domain ($L_{fMSE}$), the perceptual loss ($L_{VGG}$), and the generative adversarial loss ($L_{GEN}$). The mean square error of the image domain and frequency domain ensured the consistency of the image with the standard image in the image domain and frequency domain. The perceptual loss used the Visual Geometry Group (VGG) model [39] pre-trained on ImageNet to explore the consistency of the image in the deep features. The adversarial loss was the loss function adapted from the generative adversarial network, which D could help recover detailed features of the generated image. Therefore, the total loss function is:

$$L_{total} = \alpha L_{iMSE} + \beta L_{fMSE} + \gamma L_{VGG} + L_{GEN} \quad (11)$$

$$L_{iMSE} = \frac{\|\mathbf{x_{rec}} - \mathbf{x_{gt}}\|_2}{\|\mathbf{x_{gt}}\|_2} \quad (12)$$

$$L_{fMSE} = \|FFT(\mathbf{x_{rec}}) - FFT(\mathbf{x_{gt}})\|_2^2 \quad (13)$$

$$L_{VGG} = \|VGG(\mathbf{x_{rec}}) - VGG(\mathbf{x_{gt}})\|_2^2 \quad (14)$$

$$L_{GEN} = -\log(D_{\theta_D}(\mathbf{x_{rec}})) \quad (15)$$

where $\alpha = 60$, $\beta = 30$ and $\gamma = 0.01$ are hyperparameters. The network training took about 2 days with an NVIDIA GeForce GTX 1060 GPU, 1.6 GHz, and 6.0 GB RAM on 2400 slices (5 frames for each slice). The network was trained until the loss stabilized.

### C. Performance Evaluation

The reconstruction performance was evaluated by using Structural Similarity Index (SSIM), Peak Signal to Noise Ratio (PSNR), and Normalized Mean Square Error (NMSE). The SSIM [40] calculates the structural similarity between the reconstructed image $I_{rec}$ and the ground-truth image $I_{gt}$:

$$\text{SSIM}(\mathbf{X}, \mathbf{Y}) = \frac{(2u_{I_{rec}} u_{I_{gt}} + C_1)(2\sigma_{I_{rec} I_{gt}} + C_2)}{(u_{I_{rec}}^2 + u_{I_{gt}}^2 + C_1)(\sigma_{I_{rec}}^2 + \sigma_{I_{gt}}^2 + C_2)}, \quad (16)$$

where $\mu_{I_{rec}}$ and $\mu_{I_{gt}}$ are the average of $I_{rec}$ and $I_{gt}$, respectively; $\sigma_{I_{rec}}$ and $\sigma_{I_{gt}}$ are variance of $I_{rec}$ and $I_{gt}$, respectively; $\sigma_{I_{rec} I_{gt}}$ is the co-variance of $I_{rec}$ and $I_{gt}$, $C_1$ and $C_2$ are the two constants for stabilizing the division with weak denominator. Here we took $C_1 = 0.01^2$, $C_2 = 0.03^2$.

The PSNR uses the ratio between the maximum power of the original signal $MAX_{Igt}$ and the power of noise (the mean squared error $MSE(I_{rec}, I_{gt})$) to evaluate the reconstruction quality:

$$\text{PSNR} = 10\log_{10}(\frac{{MAX_{Igt}}^2}{MSE(I_{rec}/I_{gt})}), \quad (17)$$

The NMSE is defined as the ratio between the $l_2$ norm of error vectors and the $l_2$ norm of ground truth:

$$\text{NMSE}(I_{rec}, I_{gt}) = \frac{\|I_{rec} - I_{gt}\|_2}{\|I_{gt}\|_2}. \tag{18}$$

For DBS, the interventional feature occupies a relatively small local region. Therefore, in addition to the evaluation of the reconstruction performance of the whole image, we also evaluated with respect to a region of interest (ROI) encompassing only the intervention feature. Results were compared with a reconstruction algorithm for dynamic MRI (Golden-Angle Radial Sparse Parallel (GRASP)) [41], a similar RNN-based work [30] and a previously reported DL network for i-MRI (FbCNN) [42]. All DL methods are sufficiently trained until convergence for a fair comparison. Evaluation metrics based on the whole testing data sets (188 slices × 5 frames) were calculated.

## IV. RESULTS

The performance of the proposed method was evaluated with 5, 10, 20, and 40 spokes (TABLE I). Results demonstrated that the interventional features can be reconstructed already under 5 spokes (Fig. 4) and the reconstruction error decreased as the number of sampling spokes increased (Fig. 5). To evaluate the reconstruction error on ROIs including the interventional features, local NMSE and SSIM were also calculated (TABLE II). Compared with the global NMSE, we observed lower local NMSE values for intervention ROIs.

The reconstruction performance was also compared with GRASP and FbCNN (TABLE I). Satisfactory reconstruction results of the intervention feature could not be obtained for GRASP, CRNN, and FbCNN with the number of radial spokes less than 10 for each frame (Fig. 6). The acquisition takes approximately 6 ms for each spoke. Therefore, the general acquisition for each image is less than 100 ms when using 10 spokes, which guarantees sufficient temporal resolution.

To compare with the proposed algorithm without including the k-space data from consecutive frames, we masked the Conv-LSTM block by setting the output of Conv-LSTM block as zero, and denoted the reconstructed images from each RNN block by "RNN_masked". Under the condition of 10 spokes, the proposed method was able to reconstruct the images with NMSE of 0.201 globally and 0.170 locally, which performed better than other similar methods. The RNN_masked model could also reconstruct the interventional images with global NMSE of 0.304 and local NMSE of 0.276.

Compared with the RNN_masked model, results from the proposed methods demonstrated that utilizing the temporal coherence between different frames could help achieve better reconstruction. The RNN scheme adopted uses previously acquired images for better reconstruction of the current frame. We could have more frames for better reconstruction performance, but the reconstruction time would be prolonged.

Therefore, we also generated an extra 7 frames datasets and tested the proposed algorithm with different numbers of time frames (3, 5, and 7 frames) as input for the RNN (Fig. 7). For each case, the k-space was sampled with 10 radial spokes for each frame. Results demonstrated that reconstruction with 5 frames could provide the best trade-off between temporal resolution and image quality (TABLE III).

The latency time was estimated by considering both k-space acquisition time and the network inference time. Here, the latency time of the proposed algorithm for 3, 5, and 7 frames were 2.12 s, 3.30 s, and 4.90 s, respectively. However, the latency time could be greatly reduced by using enhanced hardware systems.

Ablation studies were further performed to investigate the role of discriminator and initializer. As shown in Fig. 8, the initializer could help incorporate the structure information from the reference image, thus improving the overall reconstruction results. The adversarial learning strategy improved the SSIM from 0.756 to 0.765 (TABLE IV), and demonstrated better reconstruction of the interventional features (ROI).

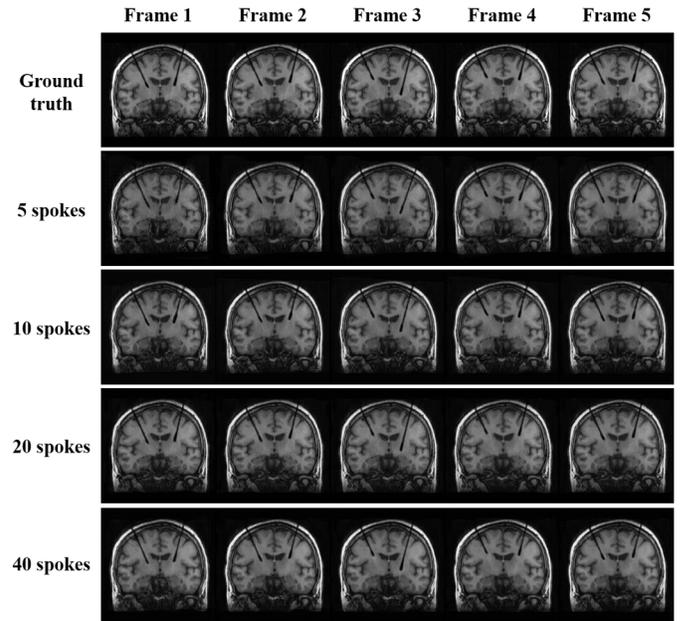

Fig. 4. Reconstructed images from the proposed method with different number of radial spokes. The five image slices were constructed from the k-space data of the five consecutive input frames.

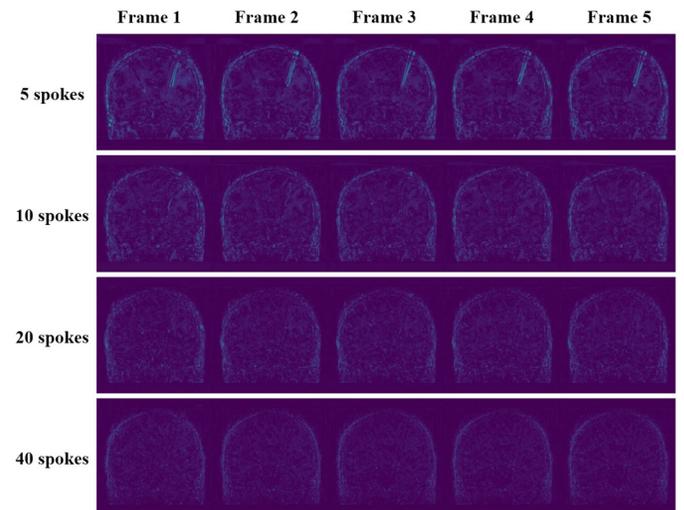

Fig. 5. The normalized error maps of Fig. 4.

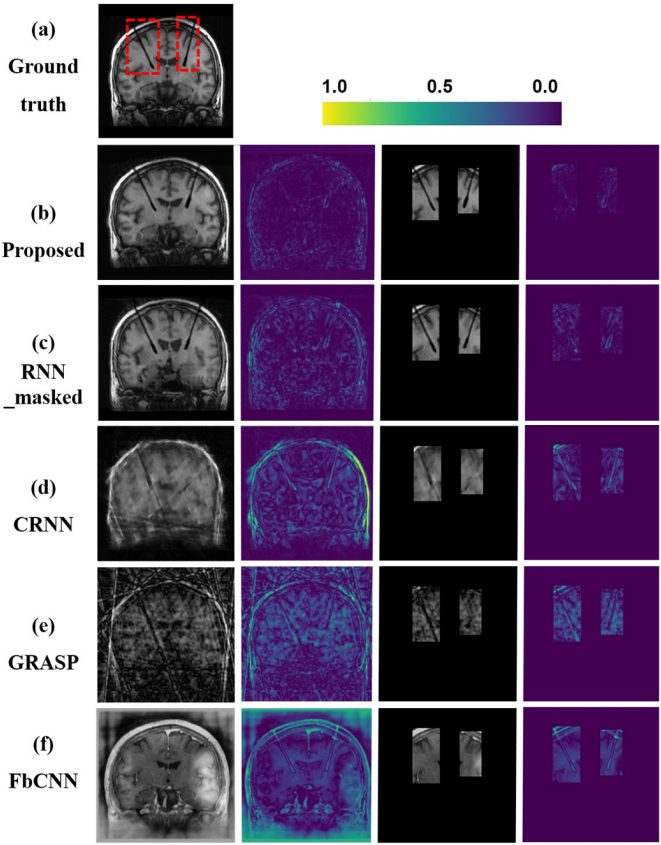

Fig. 6. A comparison of reconstruction results from different algorithms using 10 radial spokes. (a) Ground truth (b) Proposed (c) RNN_masked (d) CRNN (e) GRASP (f) FbCNN. For each method (b)-(e), the reconstructed image, global error map, reconstructed ROI, and local error map are shown.

TABLE I
A COMPARISON OF DIFFERENT RECONSTRUCTION METHODS IN
TERMS OF PEAK SIGNAL-TO-NOISE RATIO (PSNR), NORMALIZED
MEAN SQUARE ERROR (NMSE), AND STRUCTURAL SIMILARITY
(SSIM) METRICS.

| Spokes | Model | SSIM | NMSE | PSNR |
|---|---|---|---|---|
| 5 | **Proposed** | **0.674±0.061** | **0.283±0.047** | **20.578±1.686** |
| | RNN_masked | 0.463±0.037 | 0.368±0.016 | 18.171±1.060 |
| | CRNN | 0.387±0.016 | 0.437±0.041 | 16.715±0.526 |
| | GRASP | 0.247±0.010 | 0.701±0.043 | 12.590±0.704 |
| | FbCNN | 0.381±0.084 | 0.772±0.227 | 12.166±1.904 |
| 10 | **Proposed** | **0.765±0.054** | **0.201±0.035** | **23.560±1.911** |
| | RNN_masked | 0.555±0.034 | 0.304±0.018 | 19.830±1.154 |
| | CRNN | 0.517±0.025 | 0.314±0.022 | 19.552±0.859 |
| | GRASP | 0.294±0.014 | 0.569±0.044 | 14.420±0.717 |
| | FbCNN | 0.383±0.092 | 0.761±0.240 | 12.354±2.129 |
| 20 | **Proposed** | **0.769±0.041** | **0.165±0.023** | **25.212±1.537** |
| | RNN_masked | 0.662±0.025 | 0.204±0.014 | 23.317±1.112 |
| | CRNN | 0.650±0.026 | 0.237±0.018 | 22.019±1.082 |
| | GRASP | 0.374±0.017 | 0.443±0.053 | 16.619±0.720 |
| | FbCNN | 0.409±0.092 | 0.819±0.305 | 11.907±2.611 |
| 40 | **Proposed** | **0.852±0.020** | **0.109±0.010** | **28.760±1.183** |
| | RNN_masked | 0.770±0.015 | 0.138±0.007 | 26.702±0.959 |
| | CRNN | 0.761±0.021 | 0.163±0.015 | 25.261±1.367 |
| | GRASP | 0.486±0.022 | 0.347±0.052 | 18.781±0.952 |
| | FbCNN | 0.356±0.081 | 0.964±0.388 | 10.620±2.934 |

TABLE II
A COMPARISON OF THE EVALUATION METRICS FOR
RECONSTRUCTION OF INTERVENTIONAL IMAGES. THE REGION OF
INTEREST (ROI) WAS KEPT CONSISTENT THROUGHOUT THE
IMAGE FRAMES FOR EVALUATIONS OF THE LOCAL EVALUATION
METRICS.

| Spokes | Model | Local_SSIM | LOCAL_NMSE |
|---|---|---|---|
| 5 | **Proposed** | **0.748±0.070** | **0.265±0.111** |
| | RNN_masked | 0.607±0.047 | 0.381±0.131 |
| | CRNN | 0.606±0.065 | 0.418±0.161 |
| | GRASP | 0.415±0.037 | 0.643±0.058 |
| | FbCNN | 0.631±0.091 | 0.440±0.178 |
| 10 | **Proposed** | **0.813±0.058** | **0.170±0.074** |
| | RNN_masked | 0.678±0.048 | 0.276±0.088 |
| | CRNN | 0.659±0.058 | 0.293±0.078 |
| | GRASP | 0.515±0.035 | 0.498±0.059 |
| | FbCNN | 0.611±0.080 | 0.478±0.155 |
| 20 | **Proposed** | **0.845±0.041** | **0.123±0.042** |
| | RNN_masked | 0.782±0.028 | 0.166±0.049 |
| | CRNN | 0.749±0.055 | 0.221±0.055 |
| | GRASP | 0.621±0.035 | 0.382±0.068 |
| | FbCNN | 0.639±0.093 | 0.431±0.163 |
| 40 | **Proposed** | **0.901±0.020** | **0.079±0.023** |
| | RNN_masked | 0.862±0.020 | 0.112±0.030 |
| | CRNN | 0.857±0.044 | 0.144±0.027 |
| | GRASP | 0.730±0.030 | 0.306±0.061 |
| | FbCNN | 0.632±0.090 | 0.497±0.192 |

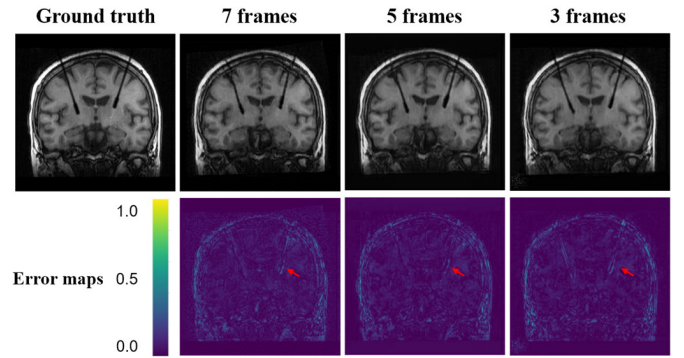

Fig. 7. A comparison of the reconstruction results of the proposed algorithm with 3, 5, and 7 consecutive frames.

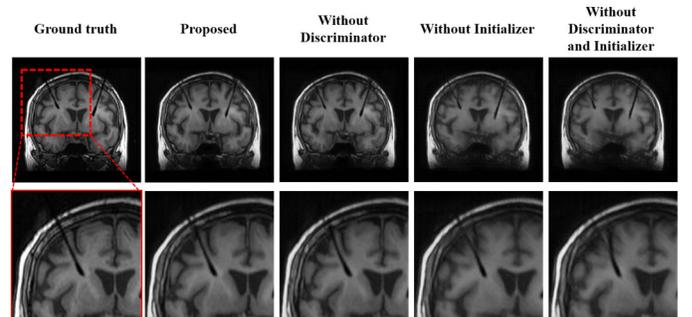

Fig. 8. A comparison of the reconstruction results of the ablation study. The ROI is extracted for better comparison.

TABLE III
A COMPARISON OF THE RECONSTRUCTION PERFORMANCE METRICS OF THE PROPOSED ALGORITHM WITH 3, 5, AND 7 CONSECUTIVE
FRAMES FOR RECONSTRUCTION

| Frames | SSIM | NMSE | PSNR |
|--------|------|------|------|
| 7 | 0.762±0.055 | 0.198±0.038 | 23.683±1.960 |
| 5 | 0.765±0.036 | 0.201±0.054 | 23.560±1.911 |
| 3 | 0.752±0.055 | 0.207±0.034 | 23.311±1.798 |

TABLE IV
A COMPARISON OF THE EVALUATION METRICS FOR ABLATION STUDY

| | PROPOSED | WITHOUT DISCRIMINATOR | WITHOUT INITIALIZER | WITHOUT DISCRIMINATOR AND INITIALIZER |
|---|---|---|---|---|
| PSNR | **23.560 ± 1.911** | 23.483 ± 1.823 | 22.606 ± 1.315 | 22.565 ± 1.327 |
| NMSE | **0.201 ± 0.054** | 0.203 ± 0.057 | 0.222 ± 0.041 | 0.223 ± 0.040 |
| SSIM | **0.765 ± 0.036** | 0.756 ± 0.035 | 0.674 ± 0.022 | 0.677 ± 0.023 |

## V. DISCUSSION

Accurate positioning of the intervention feature is a key for surgical guidance in DBS. Many models have been proposed for this purpose. For example, biomechanical models were used for the prediction of the position of interventional features [43], [44]. Another model used LSTM embedded DL approach to analyze the data from Micro Electrode Recordings (MERs) for electrode localization [45]. However, i-MRI could provide advantages with direct visualization of the intervention process, thus improving the localization of the inserted features. In this study, we proposed to use RNN for i-MRI reconstruction. The proposed method incorporates a reference image as prior information and utilizes the temporal coherence between different time frames with Conv-LSTM. The proposed method could achieve an acceleration rate up to 40 folds, which could be potentially used for real-time i-MRI.

Most of the current DL-based algorithms were proposed for dynamic imaging [46], [47]. DL networks such as Cascade CNN [24] and DAGAN [25] could reconstruct a single frame for the undersampled data. The Cascade CNN used a series of CNNs and DC layers to implement the optimization of the reconstruction algorithm. Adversarial learning was added to improve the reconstruction performance in the framework of DAGAN. The end-to-end DL methods have been validated based on clinical applications [47]. For brain interventional imaging, data acquisition and image reconstruction were carried out continuously online. Therefore, a higher acceleration rate is required with fewer sampled data. Compared with these algorithms for dynamic imaging, our method used Conv-LSTM to incorporate the prior temporal information, which is different from our previously proposed FbCNN [42]. Here, with clinical image data sets, we showed that Conv-LSTM was essential to distinguish the misalignment between the reference image and the interventional image.

RNN-based reconstruction methods have been used to explore the spatiotemporal coherence between different slices and different frames [30], [31], [48], [49]. These methods implemented the iterative reconstruction process through a cascaded CNN structure and could capture the temporal dependence. Here, the architecture used only two cascaded CNNs and an initializer for Conv-LSTM initialization.

Compared with the previous RNN based methods [30], we explored using LSTM block to incorporate the temporal information for i-MRI reconstruction. The LSTM block could help control the information flow and combine temporal information with additionally pretrained control gates. To further boost the reconstruction performance and avoid the potential over-smoothing artifacts [25], a generative adversarial framework was also adopted in the proposed algorithm. O. Jaubert et al. and Q. Lyu et al. have used Conv-LSTM for artifact suppression in cardiac imaging, but a lack of data consistency may introduce unreal features [35], [36]. In our proposed network, the DC layer was specifically designed for our radial-sampling interventional dataset. Instead of using the conventional FFT-based data consistency for hard projection, we adopted a NUFFT based gradient update step for soft projection, specifically tailored for the radially sampled k-space data reconstruction.

This work has some limitations. As a data-driven methods, the generalizability of the proposed network depends on the training datasets [22], [50], [51]. In this study, to mimic the clinical scenario of electrode implant, the training and testing datasets were based on preoperative and postoperative MRI of the 29 DBS patients. The relatively small data sets may limit the generalizability of the proposed network. Future work includes collecting and accumulating more clinical data to improve the performance of the proposed network, and carry out real-time brain i-MRI experiment for further testing and validation of the method.

In conclusion, we proposed a Conv-LSTM based RNN for reconstructing interventional images. Temporal image frames before the current reconstructed image were used. Results demonstrated the potential of the proposed network for i-MRI reconstruction. The proposed method may also be used in other interventional image reconstruction scenarios.

## ACKNOWLEDGMENT

We thank Profs. Zhi-Pei Liang, Jun Zhao, and Yiping Du for their helpful discussions.